\documentclass[lettersize,journal]{IEEEtran}
\usepackage{amsmath,amsfonts}
\usepackage{algorithmic}
\usepackage{algorithm}
\usepackage{array}
\usepackage{textcomp}
\usepackage{stfloats}
\usepackage{url}
\usepackage{verbatim}
\usepackage{graphicx}
\usepackage{cite}
\usepackage{subfigure}
\usepackage{hyperref}[colorlinks,linkcolor=blue]

\begin{document}

\title{Retinex-guided Channel-grouping based Patch Swap for Arbitrary Style Transfer}

\author{Chang Liu{\small $~^{1}$}, Yi Niu{\small $~^{1,2*}$}, \textit{Member, IEEE}, Mingming Ma{\small $~^{1}$}, Fu Li{\small $~^{1}$} and Guangming Shi{\small $~^{1}$},  \textit{Fellow, IEEE}%
\vspace{1mm}
\fontsize{10}{10}\selectfont\itshape

$^{1}$\,School of Artificial Intelligence, \\
Xidian University\\
\vspace{1mm} \fontsize{10}{10}\selectfont\rmfamily\itshape
$^{2}$\,The Pengcheng Lab, China\\

$^{*}$\,niuyi@mail.xidian.edu.cn\\

\thanks{Thanks for the support of NSFC(61875157, 61672404, 61751310), the Fundamental Research Funds for the Central Universities(JC1904).}}

%% The paper headers
%\markboth{Journal of \LaTeX\ Class Files,~Vol.~14, No.~8, August~2021}%
%{Shell \MakeLowercase{\textit{et al.}}: A Sample Article Using IEEEtran.cls for IEEE Journals}
%
%\IEEEpubid{0000--0000/00\$00.00~\copyright~2021 IEEE}
%% Remember, if you use this you must call \IEEEpubidadjcol in the second
%% column for its text to clear the IEEEpubid mark.

\maketitle

\begin{abstract} % 133 w.

%The basic principle of the patch-matching based style transfer is to substitute the patches of the content image feature maps by the closest patches from the style image feature maps. Since the finite features harvested from one single aesthetic style image are inadequate to represent the rich textures of the content natural image, existing techniques treat the full-channel style feature patches as simple signal tensors and create new style feature patches via signal-level fusion. In this paper, we propose a channel-grouping based patch swap technique to group the style feature maps into surface and texture channels, and the new features are created by the combination of these two groups, which can be regarded as a semantic-level fusion of the raw style features. Experimental results demonstrate that the proposed method outperforms the existing techniques in providing more style-consistent textures while keeping the content fidelity.

The basic principle of the patch-matching based style transfer is to substitute the patches of the content image feature maps by the closest patches from the style image feature maps. Since the finite features harvested from one single aesthetic style image are inadequate to represent the rich textures of the content natural image, existing techniques treat the full-channel style feature patches as simple signal tensors and create new style feature patches via signal-level fusion, which ignore the implicit diversities existed in style features and thus fail for generating better stylised results. In this paper, we propose a Retinex theory guided, channel-grouping based patch swap technique to solve the above challenges. Channel-grouping strategy groups the style feature maps into surface and texture channels, which prevents the winner-takes-all problem. Retinex theory based decomposition controls a more stable channel code rate generation. In addition, we provide complementary fusion and multi-scale generation strategy to prevent unexpected black area and over-stylised results respectively. Experimental results demonstrate that the proposed method outperforms the existing techniques in providing more style-consistent textures while keeping the content fidelity.

\end{abstract}

\section{Introduction}
\label{sec:intro}

\begin{figure}[h]
 \centering
   \subfigure{\includegraphics[width=0.9\linewidth]{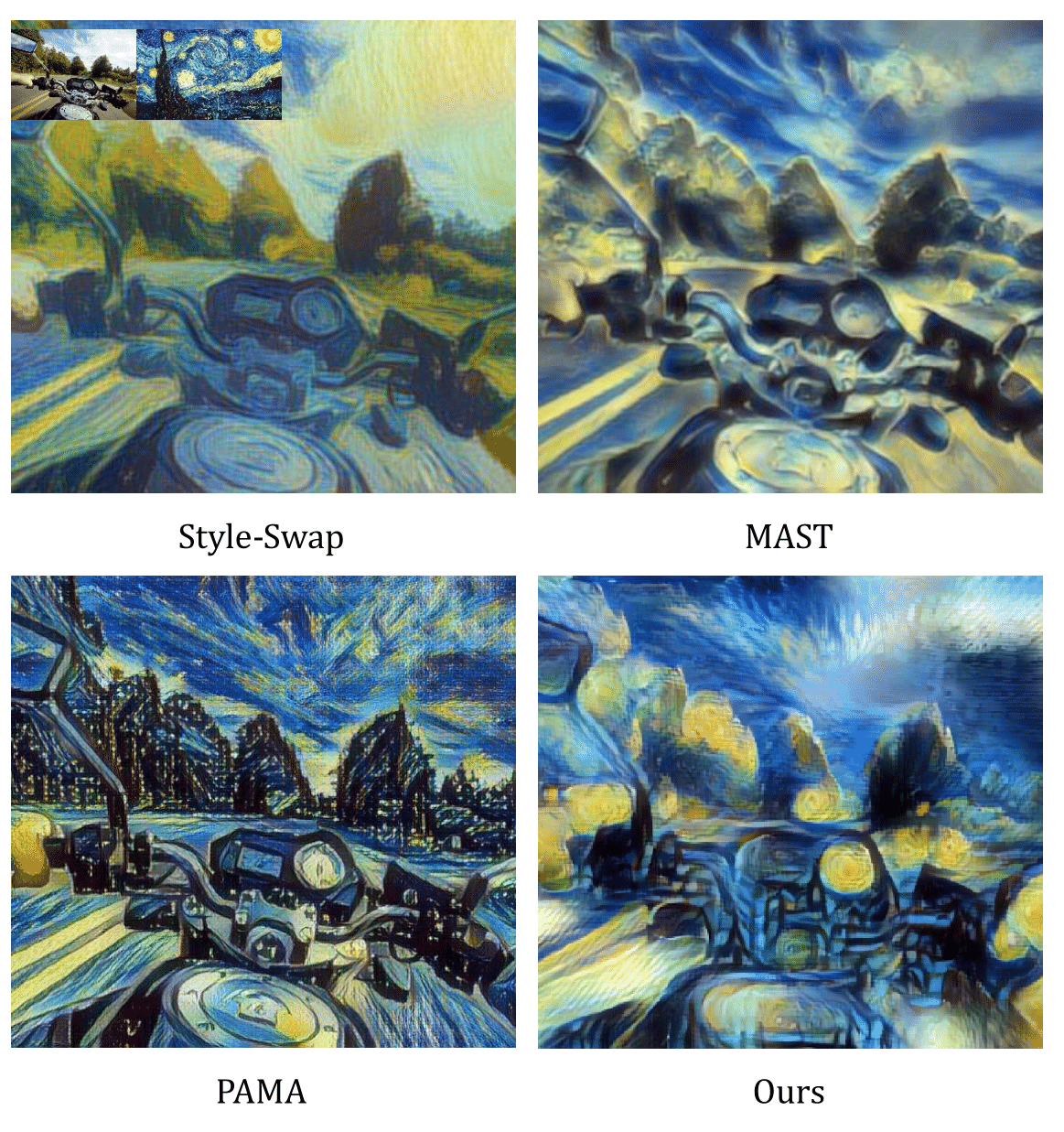}}
    \caption{Our method can extract more vivid textures from the style image.}
    \label{fig:Drawbacks}
\end{figure}

% problem很重要
The masterpiece of an artist imply great aesthetic value while few of them can survive through years. Thus, extracting and representing the style of an artwork becomes an essential task that inspires generations to inherit and revive them. Style transfer, derived from image synthesis, aiming to transfer the style from the source image to the target while keeping the original spatial structure, has been studied for years to represent and synthesize the style from art paintings to natural images.

% 分类(缩小范围)
Earlier style transfer techniques \cite{gatys2015neural,mordvintsev2015inceptionism} take one single style image and one single content image as a pair and accomplish style transfer via image optimization thus is time consuming. To improve the flexibility and efficiency, model based style transfer techniques \cite{johnson2016perceptual,ulyanov2017improved} are proposed to explicitly model the image style via offline learning and stylize the nature image via online restoration. Recently, Patch-based algorithm\cite{frigo2016split,chen2016fast,sheng2018avatar,park2019arbitrary,yao2019attention,
deng2020arbitrary,liu2021adaattn,chen2021artistic}has attracted board attention in terms of its high efficiency and \emph{arbitrary-style-per-model}\cite{jing2019neural} capability.

A milestone of the patch-matching based technique is the Style-Swap \cite{chen2016fast} which adopts an auto-encoder structure to extract features from content and style images respectively, then substitutes the content feature patches by the style feature patches, in the bottleneck layer.
A basic assumption behind Style-Swap is that the features extracted from style image are sufficient to represent the content image. This coarse assumption fails when transferring the style from a texture-free aesthetic image to the rich-textured natural images. Substituting the features directly may flatten the subtle structures and cause insufficient stylization problem, as shown in Fig .\ref{fig:Drawbacks}.

A native solution to overcome the feature insufficient problem is to create more style feature candidates to enrich the search space of patch matching.
Sheng et al. \cite{sheng2018avatar} improves Style-Swap by introducing WCT \cite{li2017universal} technique and AdaIN technique\cite{huang2017arbitrary} to modify the intensity of the swapped features according to the global statistics of the style feature maps.
Park et al. \cite{park2019arbitrary} proposes an attentional network to estimate the similarity between the ongoing content feature patches and all the style patches. Instead of using the raw style patches directly, SANet adopts an weighted average version of the style patches according to a softmax activation of the estimated similarity.
Yao et al. \cite{yao2019attention} propose to up-sample the style feature map into different scale and using multi-scale swap via a proposed self-attention auto-encoder structure.
Liu et al. \cite{liu2021adaattn} performs multi-path attantion module and utilize both shallow and deep features to generate more rich stylised results.

\begin{figure}[]
	\centering
	\subfigure[Style image.]{
		\begin{minipage}[b]{0.12\textwidth}
		\centering
		\includegraphics[width=2.2cm]{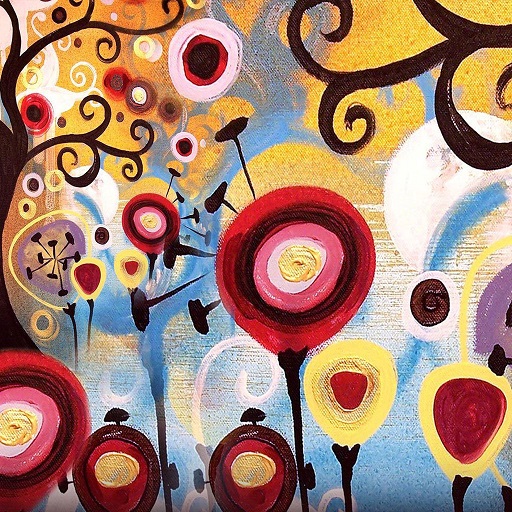}
		\label{Fig.diff-channel(a)}
		\end{minipage}
		}
	\subfigure[Surface part.]{
		\begin{minipage}[b]{0.12\textwidth}
		\centering
		\includegraphics[width=2.2cm]{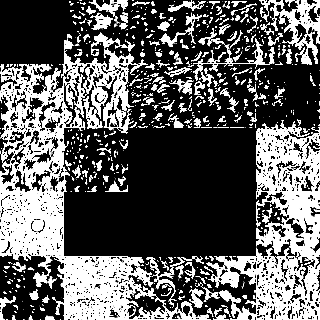}
		\label{Fig.diff-channel(b)}
		\end{minipage}
		}
	\subfigure[Texture part.]{
		\begin{minipage}[b]{0.12\textwidth}
		\centering
		\includegraphics[width=2.2cm]{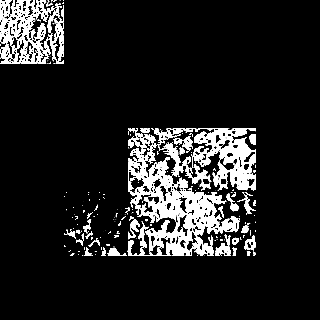}
		\label{Fig.diff-channel(c)}
		\end{minipage}
		}
\caption{Visualization of style features $F_S$ of first 25 channels extracted by encoder.}
\label{Fig.show-diff-channel}
\end{figure}

In summary, the above patch-swap-based methods explore to enrich style patches by either adopting attention modules or generating multi-scale style patches, and the basic elements performed in these methods are full-channel tensors with different spatial locations. Thus the generated style patches can be regarded as a signal-level fusion via intensity modification\cite{sheng2018avatar, li2017universal, huang2017arbitrary}, linear mapping \cite{park2019arbitrary} or multi-scale fusion\cite{yao2019attention, liu2021adaattn}.
However, as shown in Fig. \ref{Fig.show-diff-channel}, since the channels of style feature carry diverse semantic information, such as textures, colors and edge, etc, the above full-channel matching strategies suffers from two drawbacks. Firstly, from the view of numerical computation, the magnitude of different channels varies considerably and using full-channel matching may cause the \emph{winner-takes-all} problem\cite{cheng2021style}. Secondly and more importantly, taking the full-channel tensor as basic element can only generate signal-level-fused features, but fails for semantic-level-fused features. For example, full-channel-based methods cannot embed the yellow star into the motorbike's dashboard, as shown in Fig. \ref{fig:Drawbacks}.
%Secondly and more importantly, taking the full-channel tensor as basic element can only generate signal-level-fused features, but fails for semantic-level-fused features. For example, one cannot get a red oval from a red triangle and a blue oval by signal-level-fusion.

% To summarize, the basic element of the above mentioned techniques is the full-channel tensor at different spatial positions, and the generated new style features can be regarded as a signal-level fusion of these elements via intensity modification\cite{sheng2018avatar}, linear fusion \cite{park2019arbitrary} or multi-scale fusion\cite{yao2019attention}.

%Considering that different channels of the style feature map carries diverse semantic information of the style image, including textures, colors, etc, the above full-channel matching strategies suffers from two drawbacks.

% Firstly, from the view of numerical computation, the magnitude of different channels varies significantly, using the full-channel matching may cause the \emph{winner-takes-all} problem.

%Secondly and more importantly, taking the full-channel tensor as basic element can only generate signal-level-fused features, but fails for semantic-level-fused features. For example, one cannot get a red oval from a red triangle and a blue oval by signal-level-fusion.

 \begin{figure*}[h]
  \centering
\includegraphics[width=1.1\linewidth]{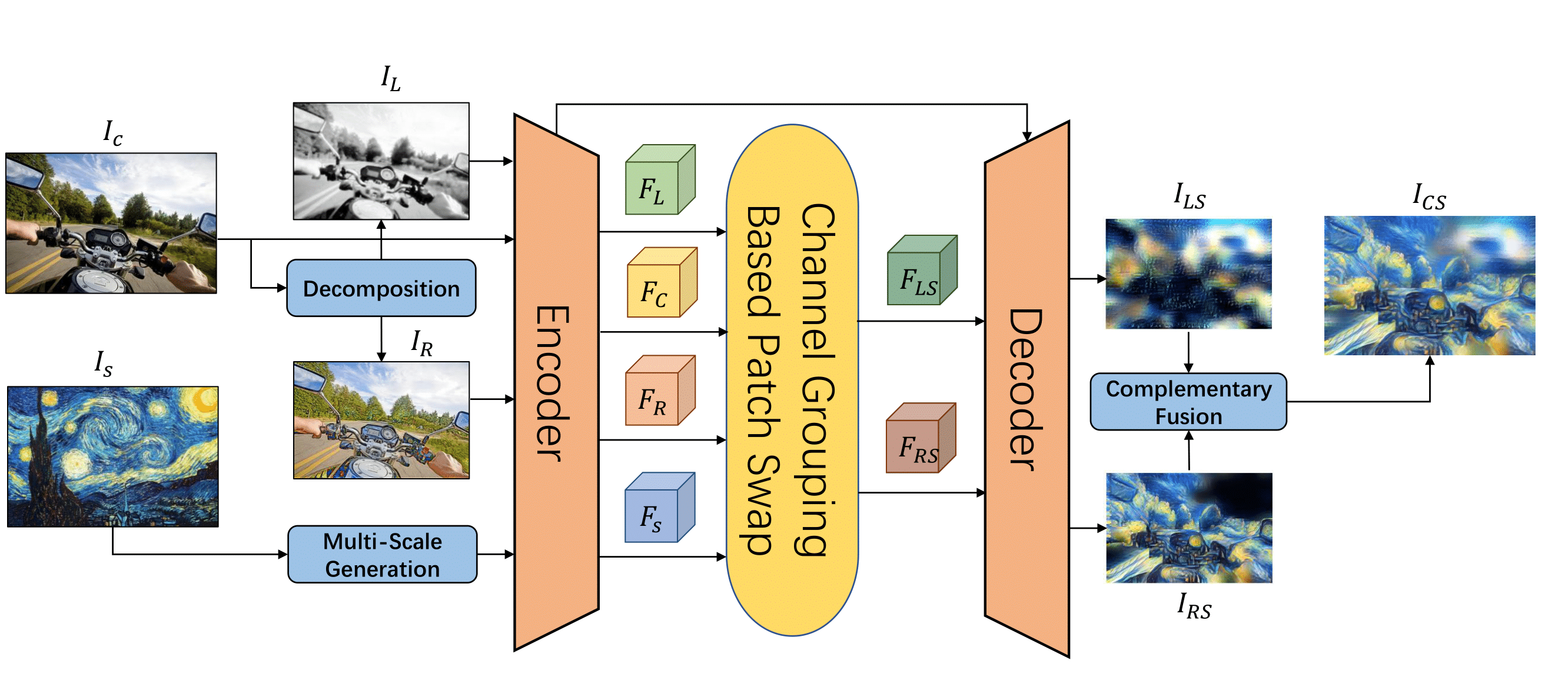}
\caption{The network architecture of the proposed R-CGPS technique. }
\label{fig:structure}
\end{figure*}

%然而，仅仅面向图像内容所进行的设计难以维持风格特征在全局与局部的一致性。为了进一步提升风格迁移性能，研究者开始利用注意力机制来进行更精细的迁移。如1,2,3

%与上述方法不同，我们

%除了基于图像自身特性的考量，越来越多的结构通过注意力机制来同时保持迁移结果在全局和局部
\noindent{ \textit{Contribution}} The goal of this paper is to perform semantic-level fusion between content and style images via the channel grouping strategy. We provide a straightforward and effective procedure to perform arbitrary style transfer. As part of this procedure, we describe a Retinex theory guided, channel-grouping based patch swap algorithm (R-CGPS). R-CGPS incorporate Retinex theory to categorize the feature map channels into texture group and surface group. Retinex decomposition helps to control a more balanced channel code rate. By different combination of the channels from the texture group and surface group respectively, new semantic features can be created and the search space can be ideally extend from $N$ to $N^2$, where $N$ denotes the spatial resolution of the style feature map. In addition, we propose to use the Unnormalized Pearson Correlation Coefficients metric to measure the similarity between content and style features for robust numerical computation. Due to the stylised results have unexpected black area, we introduce complementary fusion to embellish it. Finally, we describe a multi-scale generation strategy to rich style patches and prevent over stylised result. Experimental results demonstrate the effectiveness of proposed R-CGPS in finishing more ingenious transference and prevent over stylisation.

We note that a shorter conference version of this paper appeared in IEEE International Conference on Image Processing (2020). Our initial conference paper did not address the problem of unbalanced channel code rate and unexpected black area occurred in some stylised results. This manuscript addresses these issues and provides additional analysis on the complementary fusion and preventing over-stylised results.

\section{Related Work}

\subsection{Arbitrary Style Transfer}

Arbitrary style transfer aims at transferring arbitrary styles with one single trainable model. Inspired by the success of CNN in feature representation, researchers could focus on texture synthesis and transfer, which is categorized as parametric and non-parametric.

Parametric methods try to find a minimum number of parameter to fit any style. Ghiasi et al. \cite{google2017cin} firstly train a style prediction network that takes multiple style image as input and predicts the affine coefficients of style, then transfer content image to desired style with affine transformation. Another similar approach is proposed by Huang and Belongie \cite{huang2017arbitrary}, namely AdaIN. Instead of using network to predict style affine coefficients, AdaIN transfers the channel-wise mean and variance of content activation to match those of style. In 2020, Jing et al. \cite{jing2020din} propose dynamic instance normalization (DIN) to transform content activations with a trainable convolution kernel. In detail, a weight matrix and a bias matrix  ($1\times1$ in paper) is learned respectively from one style image, and then used to transform content activations. In summary, parametric methods usually match the statistics of content features with that of style, and then decode the transformed features to images.

Different from parametric method, non-parametric neural style transfer method transfers content images with local patch matching. The first patch-based non-parametric arbitrary style transfer method is proposed by Chen et al.\cite{chen2016fast}. They first use a pre-trained VGG19 encoder to extract content and style patches at intermediate layers, then match each content activation patch to the most similar style activation patch and swap them. The stylized result is produced by reconstructing the complete content activations with a trainable VGG decoder. Based on the patch swap strategy, Sheng et al. \cite{sheng2018avatar} propose Avatar-Net that incorporate WCT and AdaIN to project and reconstruct stylized content images. Park et al. \cite{park2019arbitrary} adopt attention mechanism to estimate the similarity between content patches and style patches. Besides, the patch swap strategy is replaced with a weighted summary of style patches in different layers. To produce more style features and enrich stroke details, Yao et al. \cite{yao2019attention} proposed an attention-aware multi-stroke style transfer model (AAMS) which consists of self-attention mechanism and multi-scale style swap and fusion modules. Jing et al. \cite{wang2021divswapper} found that by using $L_2$ normalization to style patches, one can generate more diverse stylised results. Samuth et al. \cite{samuth2022patch} explored a constrained nearest neighbor search to enforce uniform sampling of style feature patches. Jing et al. \cite{jing2022learning} proposed a graph neural network based method to perform style transfer. Different from the convolution based swap strategy, they regarded the style transfer as heterogeneous information propagation process and try to perform patch swap with graph neural network. However, as we have said before, the above mentioned techniques are based on a full-channel tensor, which ignores the importance of channels with weak value.

\subsection{Decomposition based style transfer}

There also have few works focusing on decomposing content image with different mechanisms before performing style transfer. Ding et al. \cite{ding2022deep} adopted wavelet transform to decompose content images into high frequency part and low frequency part. Qu et al. \cite{qu2021non} incorporated dictionary learning into the style transfer framework. They trained an overcomplete dictionary to map the RGB images to another high-dimension color space, and the swap was performed in the whole space (full channel). Different from the conference version and the above two methods that either separate images into low/high frequency part or perform style swap in one high-dimension space, we incorporate retinex decomposition into the framework. The style swap is performed in the texture domain and surface domain respectively. Using retinex decomposition helps to generate a more stable channel code rate, thus prevent winnner-takes-all problem (as shown in Fig. \ref{fig:retinex-distribution-decom}).

\subsection{Retinex Theory}
The retinex theory is based on the human color vision perception in real scenes. It assumes that the observed image S can be decomposed into reflectance and illumination, that is
\begin{equation}\label{retinex}
    S = R \odot L
\end{equation}
Where L and R is illumination and reflectance respectively. $\odot$ means the pixel-wise multiplication. The Reflectance represents the inherent property of captured objects, which is considered to be invariant under different lightness conditions. Illumination represents the lightness conditions from environment. The retinex theory is widely used in image enhancement field\cite{choi2008color, choi2007color, ji2009single, zotin2018fast, parihar2018study}. Besides these traditional methods, recent years, researchers try to embed retinex theory into CNN. Shen et al \cite{shen2017msr} proposed a different Gaussian convolution kernels based neural network to simulate the multi-scale retinex algorithm. Wei et al \cite{wei2018deep} proposed a two-stage network to enhance low-light images. They first de-compose the input image into reflectance and illumination. Then by denoising reflectance and enhancing illumination, they reconstruct the enhanced image. Similarly, Zhang et al\cite{zhang2019kindling} also used a two-stage strategy to enhance the low-light image. Inspired by the achievement of retinex theory in image enhancement field, we incorporate the decomposition into our style transfer method. In detail, we decompose the content image into reflectance and illumination to get texture and surface channel by encoding them with the VGG encoder. Experiment shows that replacing Gaussian blur with retinex decomposition can prevent the unbalanced selection of  channel.

\section{Proposed Method}
\label{sec:pagestyle}

In this section we discuss the details of the proposed R-CGPS technique.
In Section \ref{network structure} we introduce the network structure of R-CGPS. In Section \ref{sec:retinex decomposition}, we interpret the reason for incorporation of Retinex decomposition. In Section \ref{channel grouping}, we explain in details of the channel grouping technique based on the retinex theory. In Section \ref{loss function} we describe the Un-normalized Pearson Correlation Coefficients (UPCC) similarity metric and the loss function. In Section \ref{sec:beautify}, we talk about the reason and details about usage of complementary fusion. In Section \ref{multi-scale style}, we discuss the influence of style images with different scale on the transformation result.

\subsection{Network structure}
\label{network structure}

R-CGPS follows the channel grouping based patch swap strategy in former work. Besides, the Retinex decomposition is embedded in new framework to enforce high-frequency parts and balance the binary masks. Let $I_{C}$, $I_{S}$ and $I_{CS}$ denote the input content image, style image and the output stylized image respectively. $I_{L}$ and $I_{R}$ denote the decomposed illumination and reflectance respectively. $I_{LS}$ and $I_{RS}$ are stylized illumination and reflectance results respectively.
$F_{L}$, $F_{C}$, $F_{R}$ and $F_{S}$ denote the corresponding feature maps extracted by the encoder. $F_{LS}$ and $F_{RS}$ are swapped feature maps generated by surface swap and texture swap respectively. The main network structure is illustrated in Fig.\ref{fig:structure}.

Firstly, before accomplish channel grouping, we use Decom-Net\cite{wei2018deep} to decompose $I_{C}$ to $I_{L}$ and $I_{R}$, where $I_{L}$ represents the illumination and the $I_{R}$ represents the reflectance. Then  $I_{L}$, $I_{R}$, $I_{C}$ and $I_{S}$ are fed into the encoder (pre-trained VGG-19 up to $conv4\_1$ \cite{simonyan2014very}) to extract the feature maps $F_{L}$, $F_{R}$, $F_{C}$ and $F_{S}$, respectively. Thirdly, the above four feature maps are fed into the channel grouping sub-net to separate the surface channels and texture channels, denoted by $F_C^{sur}, F_C^{tex}, F_S^{sur}$ and $F_S^{tex}$, respectively. Then, the surface/texture swap operation is conducted to substitute the feature patches of $F_C^{sur}$ (resp. $F_C^{tex}$) by $F_S^{sur}$ (resp. $F_S^{tex}$) and perform style swap, which generates the $F_{LS}$ and $F_{RS}$. Finally,  the swapped feature maps $F_{LS}$ and $F_{RS}$ construct the $I_{LS}$ and $I_{RS}$ by the decoder and construct the final stylised result $I_{CS}$.

In our former work\cite{zhu2020channel}, a channel group based style swap strategy was proposed and a Gaussian filter was adopted to group the channel. However, there still has two questions:
\begin{itemize}
  \item How to balance the binary masks generated by channel grouping strategy.
  \item How to generate more controllable and pleased stylised results.
\end{itemize}

Besides the channel grouping strategy, the above two questions will be discussed in the following subsections respectively.

\subsection{Channel Grouping}
\label{channel grouping}
The main objective of channel grouping is to estimate two complemental binary masks $m$ and $\overline{m}$, which represent the judgement of surface or texture of each channel, respectively. Considering that $F_{R}$ can be regarded as a texture version feature map of $F_C$, our idea is to generate $m$ according to $F_{R}$ and $F_{L}$. To be specific, we use spatial Global Average Pooling to squeeze $F_{R}$ and $F_{L}$ into two 1D vector $C,\hat{C} \in {\mathbb{R}}^{n}$ respectively, where $n$ denotes the number of channels. Then $m$ is estimated via
\begin{equation}
{{m[i]=}{ \left\{ {\begin{array}{*{20}{l}}
{1 \hspace{0.8cm} \ \hat{C}[i]  > {C}[i]  }\\
{0  \hspace{0.8cm}else,}
\end{array}}\right.  }}\label{m}
\end{equation}
With the help of the channel masks, the surface $F_{}^{sur}$and texture $F_{}^{tex}$ can be simply computed by
\begin{align}
F_{}^{sur}=F_{} \cdot m \hspace{0.5cm}
F_{}^{tex}=F_{} \cdot \overline{m},
\end{align}
where $\overline{m}$ is the binary reverse of $m$. In this way, the content feature map $F_c$ and style feature map $F_s$ are grouped into  $F_c^{sur}$ and $F_c^{tex}$ (resp. $F_s^{sur}$ and $F_s^{tex}$).
\begin{figure}[htpb]%tpb
%\centering
\includegraphics[width=1\linewidth]{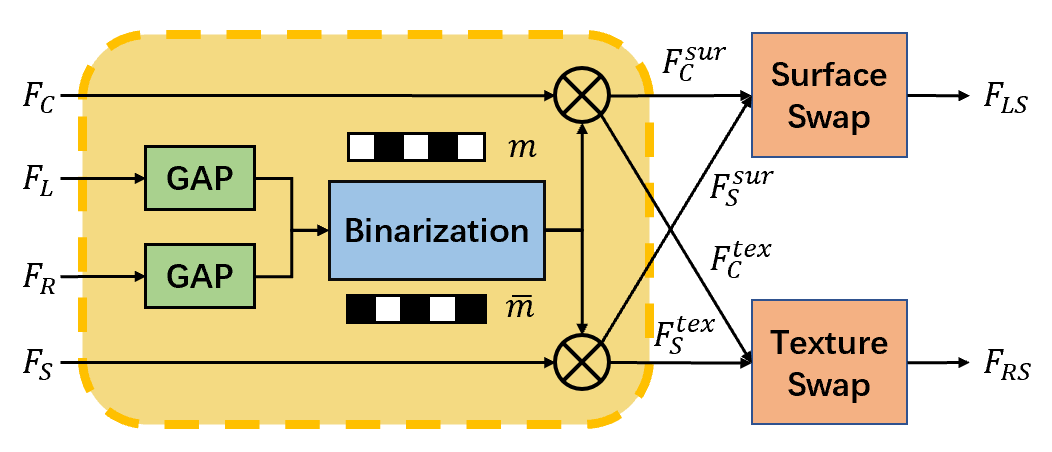}
\caption{The channel grouping sub-net.}
\label{fig:sm}
\end{figure}
Comparing with the original full-channel feature swap strategy\cite{chen2016fast} which can only match a content feature patch along $N$ style patches, channel grouping can expand the search space from $N$ to $N^2$ with the combination of $F_s^{sur}$ and $F_s^{tex}$, and accomplish a semantic-level feature fusion instead of signal-level fusion.

\subsection{Similarity Metric and Loss Function}
\label{loss function}
  % can be created  denoted by the superscript $sur$ and $tex$ respectively.The only difference is the feature swap operation.
%
%As we mentioned before, instead of swapping the full-channel features, we grouping the raw feature map into texture group and surface group.
Another important issue for patch swap is the metric of the semantic similarity between content and style features. Instead of using the widely-adopted  Normalized Cross Correlation metric which only characterize the cosine distance between two tensors, we propose to use the Un-normalized Pearson Correlation Coefficients metric (UPCC) which characterize both the cosine distance and the intensity distance between two patches.

To be specific, for each spatial location $i$ , the content patch $\phi_{i}\left({F}_{c}\right)$ searches through all style patches $ \phi_{j}\left({F}_{s}\right)$ and swap with its nearest-matching style patch under the UPCC metric as
\begin{align}
 \phi_{i}\left({F}_{cs}\right )\leftarrow \mathop{argmax}\limits_{{  \phi_{j}\left({F}_{s}\right)  }}   \left\langle   \phi_{i}\left(\mathring{F}_{c}\right) \cdot  \phi_{j}\left(\mathring{F}_{s}\right) \right\rangle,\text{ }
\end{align}
where $\mathring{F}$ is the feature map with channel-wise mean subtracted.
And the above process can be accelerated with a series of convolution and deconvolution operations\cite{chen2016fast,sheng2018avatar} as
\begin{align}
{F_{cs}=\Phi\left({F}_{s}\right)\text{  } { \ast^\mathsf{T} }\text{  } {\rm B}  \left(   \mathring{F}_{c}    \ast  \Phi\left(\mathring{F}_{s}\right)  \right   ),}
\end{align}
where $ \Phi\left({F}_{s}\right) $ is the style kernel stacked with all style patches $ \phi_{j}\left({F}_{s}\right) $  along channel axis.  $B(\cdot)$ is the binarizing operation with the maximum value as 1 and the rest as 0 along channel.  $\ast$  and $\ast^\mathsf{T}$ denote convolution and deconvolution operation respectively.

With the swapped feature map $F_{cs}$,  the parametric decoder reconstructs it to the stylized image.  To train the parametric decoder, the loss function\cite{chen2016fast}\cite{sheng2018avatar} is defined as
%The pre-trained encoder embeds the content and style image to feature space, in which the stylization proceeds.
\begin{align}
\ell = { \left\Vert {{I-\mathop{{I}}\limits^{ \sim }}} \right\Vert } _{{2}}^{2} +
 { \mathop{\sum}\limits_{i \in L} \omega_{i}\left\Vert {{Enc^{i}(I)-\mathop{{Enc^{i}(\mathop{I}\limits^{ \sim })}}}} \right\Vert_{{2}}^{2}}
 +\ell_{tv},
\end{align}
where the $\mathop{I}\limits^{ \sim }$ is the reconstructed image. Here the first item is the Euclidean distance between the input and reconstructed image and the second is perceptual losses\cite{johnson2016perceptual} of the layers set $ L=\lbrace conv1\_1, conv2\_1, conv3\_1, conv4\_1  \rbrace$,  weighted by $\omega_{i}$ respectively. $\ell_{tv}$ is the total variation loss\cite{rudin1992tv} to enhance output smoothness.

\subsection{Image Decomposition}
\label{sec:retinex decomposition}

\begin{figure}[htbp!]
%\centerline{\text{\quad Inputs  \qquad  Style-Swap  \qquad    AAMS  \qquad  \quad   Ours} }
%\centerline{\includegraphics[width=1\linewidth]{result/comparison/comparisons.jpg}}
\centerline{\includegraphics[width=0.6\linewidth]{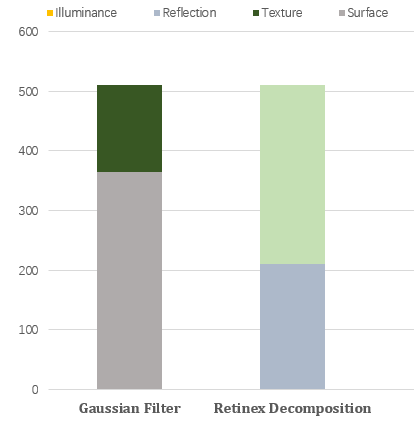}}

\caption{Comparison of average binarization code of 25 images generated by different methods. Compared with Gaussian filter, the Retinex decomposition controls a more balanced channel code rate.}
\label{fig:gfvsretinex}
\end{figure}

In former work\cite{zhu2020channel}, a Gaussian low pass filter was adopted to extract texture part and surface part in an image. Ideally, these two feature maps jointly separate the original image's channel. However, sometimes the texture part can't prove to win the comparison with texture part, which may cause an unbalanced binarization code rate, as shown in Fig. \ref{fig:gfvsretinex}. Before performing channel grouping, the $F_{C}$ and $\hat{F}_{C}$ are increased to 512 channels with quadruple downsampling of scale. By using Global Average Pooling, the feature maps are compressed to 1D vector with length of 512. Since the value of surface code $\hat{C}$ always larger than texture code $C$, it may seriously weaken the channel grouping's role.

To increase the number of texture code and stabilize the ratio of surface and texture in binarization code, we replace the Gaussian filter with Retinex decomposition. Different from Gaussian filter, Retinex decomposition decomposes images into illumination and reflectance. Detailly, reflectance describes the intrinsic properties of the object and illumination more focus on the lightness from environment. Thus, reflectance contains more details about texture and illumination is a bit smooth, which corresponds to texture and surface respectively. A more detailed ablation study can be found in Sec.\ref{sec:ablation:retinex decom}.
%In this paper, we use Decom-Net\cite{wei2018deep} to estimate reflectance and illumination directly from source content image. The Decom-Net uses a data-driven way

\subsection{Complementary Fusion}
\label{sec:beautify}

\begin{figure}[]

\centerline{\includegraphics[width=1\linewidth]{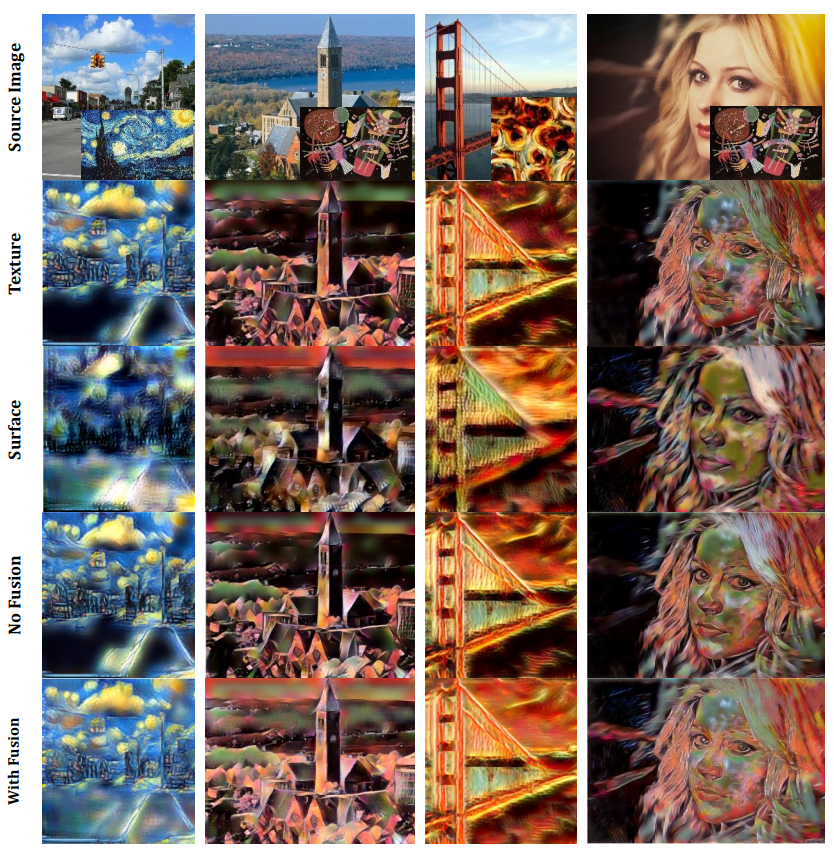}}

\caption{Ablation study about complementary fusion. Using weighted sum strategy can help to repair the unexpected black area or blank area in texture field.}
\label{fig:surface-texture}
\end{figure}

Different from global-statistic-based arbitrary style transfer methods, the patch-swap-based methods usually generate more richer stylised results while its details may be un-controlled, as shown in \ref{fig:surface-texture}. By using channel grouping based patch swap strategy, the yellow star in starry night is successfully embedded in the dashboard of motorbike, which is satisfied. However, there still has large black area appeared in the sky (in Fig. \ref{fig:structure}) and road (in Fig. \ref{fig:surface-texture}). In former work, one has to manually decide the global weight value between texture features and surface features to supress unexpected area. In the above-mentioned section, we incoporate the Retinex theory into our framework to improve the selected number of surface channel. Except this, we found that stylised surface and texture focus on different parts. For example, the second row and third row in Fig. \ref{fig:surface-texture} represent the stylised result only with texture and surface respectively. Obviously, the stylised texture results more clear and prefer the foreground while the surface focus on background and a bit fuzzy. These different results help us has a better understanding about our channel grouping strategy, that is texture plays a major role in style transfer, while surface assists to restore the area that is ignored by texture. Now we talk about implementation details.

Consider the complementary relationship between texture and surface, we convert the pixel value of texture to complementary weights by using the following equation:
\begin{equation}\label{normalize}
  cw_{idx} = 1-x_{idx} / Max , idx\in W\times H
\end{equation}
Where $x_{idx}$ is the pixel value of an image in position $idx$, thus $cw_{idx}$ is the calculated complementary weight value in position idx. $Max$ is a constant value that set to 255, $W$ and $H$ are weight and height of the image respectively. Before the multiplication, there still need one more step.

\begin{figure}[]

\centerline{\includegraphics[width=1\linewidth]{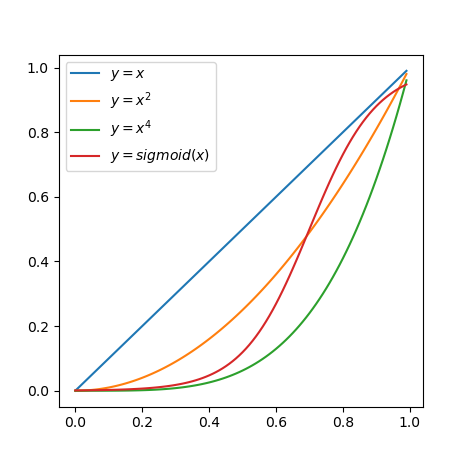}}

\caption{Compared with linear and other non-linear functions, sigmoid curve has better control over the different ranges.}
\label{fig:complementary}
\end{figure}

The value of complementary weight reflect the degree of repair required. The larger the weight value, the more repair is needed. To repair the texture and keep details, we transform complementary weight with a modified version of sigmoid function:
\begin{equation}\label{sigmoid}
%s = 1 / (1 + np.exp(-x*10+7))
  \hat{cw}_{idx}= 1/(1+\exp(-cw_{idx}*\delta)+\epsilon)
\end{equation}
where $\delta$ and $\epsilon$ are constant value. As illustrated in Fig. \ref{fig:complementary}, the sigmoid curve remains tiny value within the range [0,0.6], but it rises once the weight larger than 0.6, and the rate of increase is much higher than other non-linear curves. Thus, the complementary fusion can be finished with following equation:
\begin{equation}\label{com-beautify}
  I_{CS}=I_{RS}+\hat{CW}\circ I_{LS}
\end{equation}
Where $\hat{CW}$ is consist of $\hat{cw}_{idx}$ and has the same size with $I_{RS}$ and $I_{LS}$, $\circ$ means the pixel-wise multiplication.

\subsection{Multi-scale Style Swap}
\label{multi-scale style}
\begin{figure}[]
	\centering
	\subfigure[Original.]{
		\begin{minipage}[b]{0.1\textwidth}
		\centering
		\includegraphics[width=2cm]{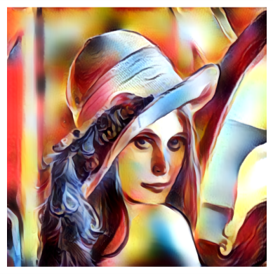}
		\label{Fig.overstylise(a)}
		\end{minipage}
		}
	\subfigure[Half only.]{
		\begin{minipage}[b]{0.1\textwidth}
		\centering
		\includegraphics[width=2cm]{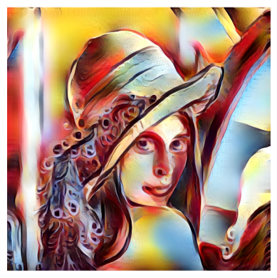}
		\label{Fig.overstylise(b)}
		\end{minipage}
		}
	\subfigure[Adding half.]{
		\begin{minipage}[b]{0.1\textwidth}
		\centering
		\includegraphics[width=2cm]{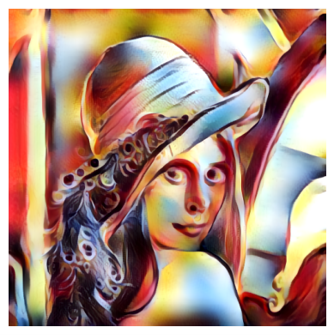}
		\label{Fig.overstylise(c)}
		\end{minipage}
		}
    \subfigure[Adding two thirds.]{
		\begin{minipage}[b]{0.1\textwidth}
		\centering
		\includegraphics[width=2cm]{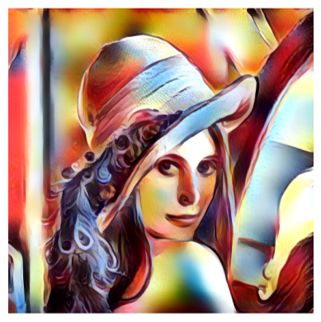}
		\label{Fig.overstylise(d)}
		\end{minipage}
		}
\caption{Different sizes of style image can produce different results. Half and two thirds means the size of style image is half and two thirds of original respectively.}
\label{Fig.overstylise}
\end{figure}

The style images with different sizes often result in different visual effects. Generally, more rich the style patches, more pleased the result. Thus researchers often choice a combination of style images with size of half or quarter. However, such combination may produce over stylised result, as shown in Fig.\ref{Fig.overstylise(c)}. To rich the style patches and prevent the over stylise results, we adopt a simple yet strategy, that is to combine the style patches with size of reducing half and third. Fig.\ref{Fig.overstylise(d)} shows that this stragety performs a balanced stylised result.

\begin{figure}[h]
 \centerline{\includegraphics[width=1\linewidth]{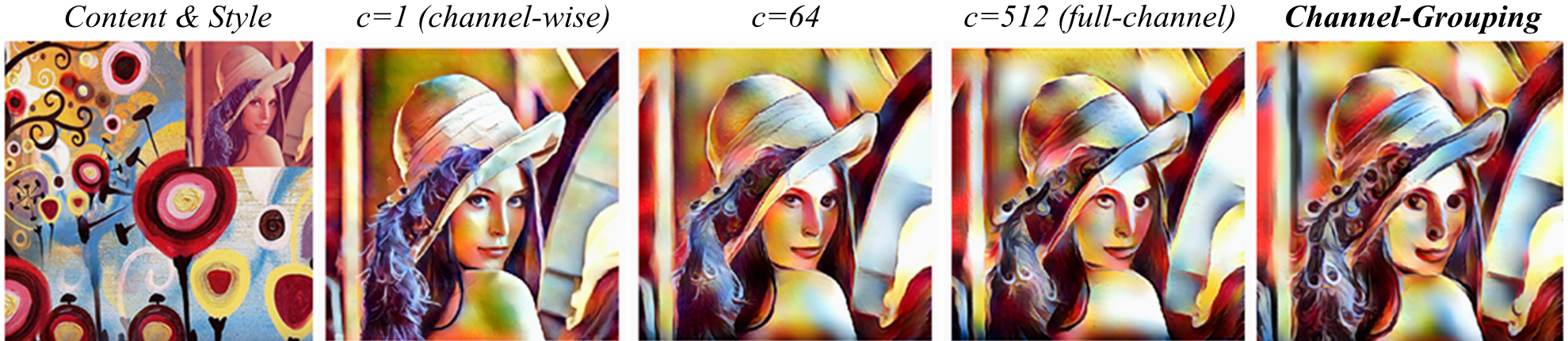}}
\caption{The comparison of channel grouping and fixed channel swap.}
\label{fig:dw}
\end{figure}

\section{Experiments and Results}
%\label{sec:typestyle}
\subsection{Implementation Details}
We train the decoder on the MS-COCO\cite{lin2014microsoft} data set about 5 epochs with a batch size of 16. To better reconstruct stylized activations,  we enhance it with Painter-By-Number data set\cite{pbn} which consists of various artworks. Adam optimizer is applied with a small learning rate of 0.0001 to avoid noised over-fitting output. In training phase, an extra encoder is connected to the output to compute the perceptual losses. Skip-connections are added to enhance stylization\cite{huang2017arbitrary}. The patch size is set to $3 \times 3$ for both groups in our experiment. To decompose content image $I_c$ to $I_L$ and $I_R$, the Decom-Net\cite{wei2018deep} is adopted in our method.

\subsection{Comparisons}
\begin{figure*}[h]
%\centerline{\text{\quad Inputs  \qquad  Style-Swap  \qquad    AAMS  \qquad  \quad   Ours} }
%\centerline{\includegraphics[width=1\linewidth]{result/comparison/comparisons.jpg}}
\centerline{\includegraphics[width=0.9\linewidth]{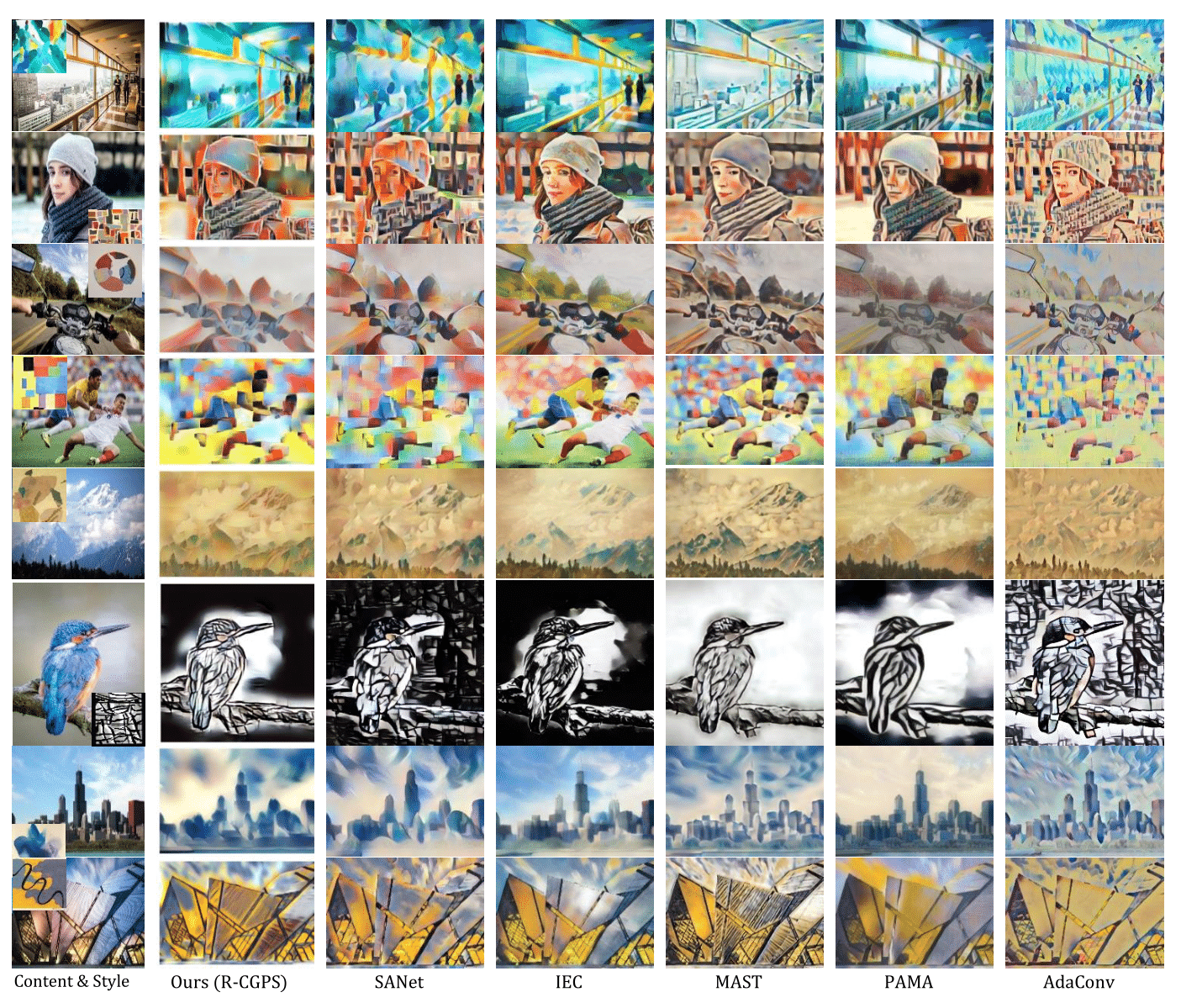}}

\caption{Our method produce comparable stylised results when compared with other methods. Channel-grouping strategy controls the balance between textures and surfaces.}
\label{fig:compare}
\end{figure*}

To validate our method, we compare our results with SANet\cite{park2019arbitrary}, IEC\cite{chen2021artistic}, MAST\cite{huo2021manifold}, PAMA\cite{luo2022progressive} and AdaConv\cite{chandran2021adaptive}, as shown in Fig.\ref{fig:compare}.

SANet tends to over-stylise images, as we can see, the detailed structure of soccer player and buildings are lost. IEC, MAST and PAMA preserve too much content structure, which cause under-stylised results. AdaConv learns to predict conv kernels in decoding layers to produce stylised results. The observed results show that the AdaConv prefers to produce much stylised textures even in smooth background.

In contrast, taking advantage of the channel grouping and semantic-level fusion strategy,  the proposed approach is able to extract more intrinsic style textures while maintain the original content in most cases.

\subsection{Ablation Study}

%\textbf{Retinex Decomposition}
%In former work, we adopt Gaussian filter to generate textures and surfaces from source images. However, the produced surfaces may much stronger than textures, which causes an unbalanced number of channel code, as shown in \ref{fig:gfvsretinex}. with Gaussian filter, the number of channels belong to surface is 511, which is much larger than that belong to texture. After changing Gaussian filter to Retinex decomposition, the surface channels are depressed to 142 and texture channels increase to 369. This shows that the Retinex decomposition can help to control a more balanced channel code for content images.

\textbf{Retinex Decomposition}
\label{sec:ablation:retinex decom}

\begin{figure}[h]
%\centerline{\text{\quad Inputs  \qquad  Style-Swap  \qquad    AAMS  \qquad  \quad   Ours} }
%\centerline{\includegraphics[width=1\linewidth]{result/comparison/comparisons.jpg}}
\centerline{\includegraphics[width=01\linewidth]{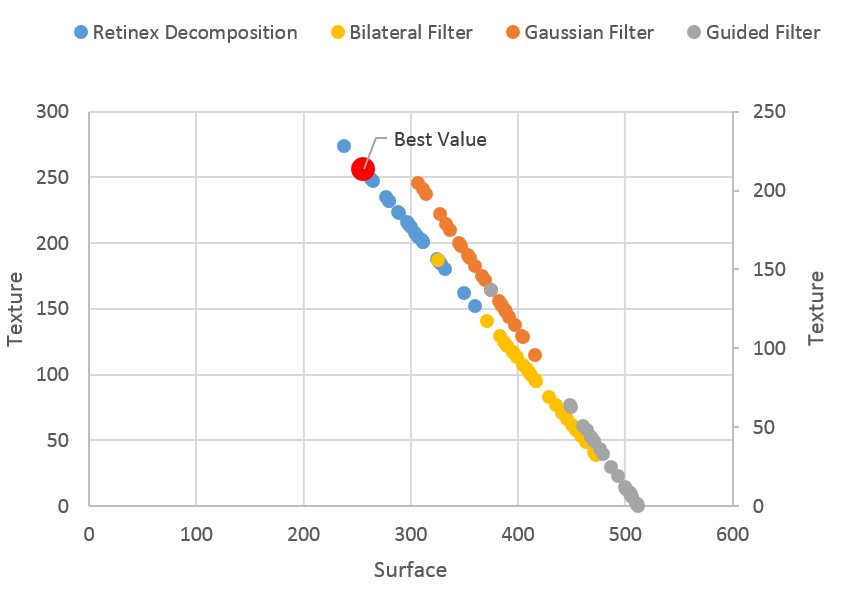}}

\caption{Comparison of rate of binarization masks generated by different methods. Retinex and Bilateral filter are drawn in primary axis, the Gaussian and Guided filter are drawn in secondary axis. Obviously, Retinex decompositin controls a more stable rate.}
\label{fig:retinex-distribution-decom}
\end{figure}

In Sec. \ref{sec:retinex decomposition}, we have noticed that the Gaussian filter may produce unbalanced channel codes rate thus adopt Retinex theory to solve this problem. To validate the effectiveness of Retinex decompositon, we compare it with three commonly used filters, as shown in Fig. \ref{fig:retinex-distribution-decom}. Here each method calculates binary masks of 25 content images and draw them as 2-D points. Consider feature maps have 512 channels, we regard the point with coordinate (256,256) as the best value. It can be seen that the masks generated by Retinex decomposition have more balanced results.

\textbf{Channel Grouping}
To validate the effectiveness of channel grouping, we directly group the feature map in neighbour channels and apply group-wise swap without the channel mask. With the number of channels in each group decreasing,  the outputs are much similar with the content image. The extreme case occurs in channel-wise swap,  as the Fig. \ref{fig:dw} shows. The stylized image is extremely similar with the content image except some color shifts while the full-channel matching fails to produce sufficient textures.  We notice that textures only exist in correct group of channels.  If those channels tangle with each other,  texture is overwhelmed in feature maps and hard to extracted.  Therefore, appropriate channel grouping plays a key role in produce implicit stylized features.

\begin{figure}[h]
%\centerline{\text{\quad Inputs  \qquad  Style-Swap  \qquad    AAMS  \qquad  \quad   Ours} }
%\centerline{\includegraphics[width=1\linewidth]{result/comparison/comparisons.jpg}}
\centerline{\includegraphics[width=01\linewidth]{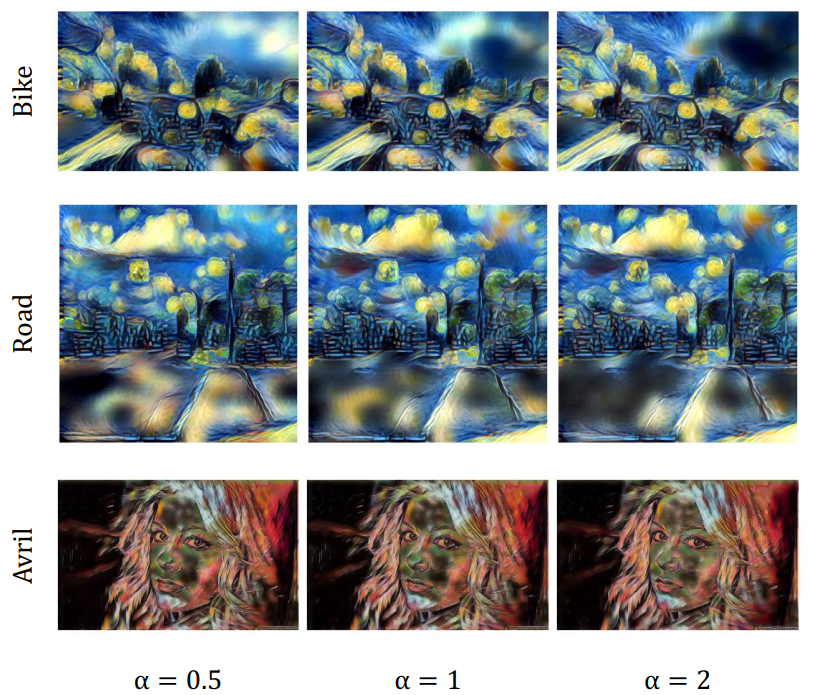}}

\caption{Controlling range of relaxtion factor is not an effective way to suppress black area.}
\label{fig:relaxtion-factor}
\end{figure}

\textbf{Complementary Fusion}
Complementary fusion is designed to suppress the unexpected black area in stylised results. In \cite{zhu2020channel}, Zhu et al added a relaxation factor $\alpha$ to control stylised results. As shown in Fig. \ref{fig:relaxtion-factor}, by controlling value of $\alpha$, the black area in "Bike" is successfully solved. However, in other two images, the black area still exists, e.g. street in "Road" and face in "Avril". Thus, in this paper, we propose a complementary fusion strategy to directly suppress these black area. The results shown in Fig. \ref{fig:surface-texture} prove that the proposed strategy can solve it well.

\textbf{User Study}
We perform user study between the proposed method and the above five approaches, as shown in Table. \ref{tab:userStudy}.
In the study, one sample consists of a content image, a style image, and six corresponding stylization results generated by the eight methods. We use 20 content images and 20 style
images to generate 400 samples and randomly draw 15 samples for each user.
For each sample, users are asked to choose the best stylization result based on their subject judgement. The result demonstrates the effectiveness of the proposed method.

\begin{table}[h]
\centering
\label{tab:userStudy}
\caption{User Study}
\resizebox{0.8\linewidth}{!}{%
\begin{tabular}{cccccc}
\hline
Ours & IEC & SANet & AdaConv & MAST & PAMA \\ \hline
\textbf{19.2$\%$} & 18.3$\%$ & 16.7$\%$ & 15.8$\%$ & 15.0$\%$ & 15.0$\%$  \\ \hline

\end{tabular}%
}

\label{tab:my-table}
\end{table}

\textbf{Efficiency}
The execution time of the above arbitrary style transfer methods is listed in Table. \ref{tab:timeCompare}. We compared our method with SANet, MAST, IEC, PAMA and AdaConv under RTX2080Ti GPU device.
Since our method has two stages, we test the execution time of each part respectively.

We compare the average execution time of six methods under two different images with size of 512 and 1024 respectively.
It can be seen that our R-CGPS cost less time than AdaConv, MAST and PAMA, and still competitive with SANet and IEC.

%It can be observed that although the ideal search space is extended from $N$ to $N^2$, the execution time of CGPS is less than two times of the original Style-Swap \cite{chen2016fast}, and much faster than AAMS method\cite{yao2019attention}.

\begin{table}[h]
\centering
\label{tab:timeCompare}
\caption{Execution time comparison(in seconds).}
\resizebox{0.8\linewidth}{!}{%
\begin{tabular}{lcc}
\hline
Method            & 512 px   & 1024 px \\ \hline
SANet              & 0.43          & 1.22           \\
IEC              & 0.47          & 1.34           \\
AdaConv           & 3.80          & 4.10           \\
MAST              & 2.33          & 3.34           \\
PAMA           & 0.77          & 1.75           \\ \hline
Ours (Retinex)    & 0.59          & 1.42           \\
Ours (CGPS)       & 0.12          & 0.25           \\ \hline
\textbf{Ours (total)} & \textbf{0.71} & \textbf{1.67}   \\ \hline
\end{tabular}%
}

\label{tab:my-table}
\end{table}

%\begin{table}[h]
%\centering
%\label{tab:timeCompare}
%\caption{Execution time comparison(in seconds).}
%\resizebox{0.8\linewidth}{!}{%
%\begin{tabular}{lcc}
%\hline
%Method            & 512 px   & 1024 px \\ \hline
%AAMS              & 1.20          & 1.79           \\
%AdaConv           & 3.80          & 4.10           \\ \hline
%Ours (Retinex)    & 0.59          & 1.42           \\
%Ours (CGPS)       & 0.12          & 0.25           \\ \hline
%\textbf{Ours (total)} & \textbf{0.71} & \textbf{1.67}   \\ \hline
%\end{tabular}%
%}
%
%\label{tab:my-table}
%\end{table}

\section{Conclusion}
\label{sec:conclusion}
In this paper we propose a Retinex theory guided, channel-grouping based patch swap technique for arbitrary style transfer. Retinex decomposition helps generate more balanced binarization masks, channel-grouping strategy enlarges the original search space for swapping style patches, multi-scale style images and complementary fusion help improve visual qualities and suppress unexpected black areas of stylised results. The relative experiments show the proposed method generates more appropriate stylised results with competitive time consuming. We believe that the proposed channel-grouping strategy can be easily extended to other tasks that are related to computer vision. In future, we plan to apply such a image decomposition and channel grouping strategies to image compression field.

Note that this paper is an extension of conference version appeared in International Conference on Image Processing (2020)\cite{zhu2020channel}. Based on the former work, we replace the Gaussian filter with Retinex decomposition, generate multi-scale style images and adopt complementary fusion strategy to suppress black areas that the initial conference paper did not addressed. One drawback of our method is that one need to manually confirm whether the stylised result should be performed complementary fusion. For this work, designing a proper image aesthetic evaluation mechanism to automatically perform decision also is a topic suitable for future work.

%In this paper we propose a Retinex guided channel-grouping based patch swap technique for arbitrary style transfer. It can be observed that the channel grouping and semantic-level fusion can significantly improve the visual quality of the stylized image than the full-channel signal-level fusion. However, this paper is only a proof-of-concept of the channel grouping strategy. In our future works, we will investigate the learnable channel grouping network to replace the simple Gaussian low pass filter and binarization.
\bibliographystyle{IEEEbib}
\bibliography{refs}

\end{document}